# BayesSeg: A Bayesian Optimization Framework for State Segmentation of Electricity Consumption Time Series

Zhenya Zhang [1], Wendi Zhu [2], Ping Wang [2], Hongmei Cheng [3],* and Shuguang Zhang [4]
[1]Anhui Province Key Laboratory of Intelligent Building & Building Energy Saving, Anhui Jianzhu University, Hefei 230022, China; zzychm@ustc.edu.cn
[2]School of Electronic and Information Engineering, Anhui Jianzhu University, Hefei 230022, China; zhuwendi@stu.ahjzu.edu.cn(W.W.); 316123054@qq.com(P.W.)
[3]School of Economics and Management, Anhui Jianzhu University, Hefei 230022, China; hmcheng@mail.ustc.edu.cn
[4]Department of Statistics and Finance, School of Management, University of Science and Technology of China, Hefei 230026, China; sgzhang@ustc.edu.cn
*Correspondence: hmcheng@mail.ustc.edu.cn

**Abstract:** In Non-Intrusive Load Monitoring (NILM), adaptive segmentation of electricity consumption time series is critical for appliance recognition. However, prevailing methods face challenges including heuristic parameter tuning, boundary sensitivity, and metric saturation. This paper proposes BayesSeg, a unified framework integrating time-series segmentation, multidimensional evaluation, and automatic parameter optimization. The segmentation layer employs a dual steady-state criterion based on the tail value and mean of preceding subsequences, combined with a sequential extraction and complement-set parsing strategy, to achieve precise unsupervised partitioning of steady-state and transition-state segments. The evaluation layer maps segmentation results to binary state sequences and formulates a composite metric integrating an event-level F1 score (event_F1) with Normalized Mutual Information (NMI). The event_F1 quantifies switching-event precision and recall via tolerance matching, while NMI captures global structural consistency, jointly overcoming the boundary sensitivity and limited discriminability of point-wise metrics. In the optimization layer, the composite score serves as the objective function for Bayesian optimization, which constructs a TPE surrogate model for efficient global parameter-space exploration. Experiments on the SustDataED2 dataset demonstrate that Bayesian optimization requires only ~100 objective evaluations to locate a parameter region within 0.35% deviation of the exhaustive grid-search optimum. The framework achieves a weighted composite score of 0.7149 and an event_F1 of 0.9340 while reducing optimization latency from ~5300 seconds to under 1 second, a speedup exceeding 5700x. BayesSeg automates segmentation configuration and provides a scalable, efficient solution for time-series analysis in NILM and related domains.

## 1. Introduction

In building energy management and demand-side response, modern electrical monitoring systems generate high-resolution electricity consumption time series from sensors such as smart meters, smart circuit breakers, and smart sockets [1-3, 16-20]. These time series capture dynamic variations in electrical parameters, such as current and power, and implicitly encode information about appliance operating states and their transitions. In particular, appliance startups and shutdowns induce significant fluctuations in the supply circuit, whereas during steady-state operation these physical quantities vary more smoothly [4-5]. Partitioning electricity consumption time series into distinct steady-state and transition-state segments based on load operating states is a fundamental prerequisite for Non-Intrusive Load Monitoring (NILM), appliance state identification, and downstream energy analytics tasks [3]. In recent years, NILM research has advanced rapidly with deep learning architectures [16, 18, 20], transfer learning strategies [17], and real-time event detection methods [5, 19], but accurate and adaptive segmentation of the underlying time series remains a common requirement across these approaches.

In time-series analysis, common segmentation approaches include change-point detection, boundary detection, and state detection [6-9, 21-26]. Specifically, multivariate regression combined with machine learning enables adaptive change-point detection in financial time series [7]; multi-task learning frameworks jointly perform activity recognition and boundary localization on wearable sensor data [8]; and self-supervised methods such as the Classification Label Profile approach identify latent states and their transitions without labeled data [9]. Although these methods have shown varying degrees of success across domains, they are generally sensitive to parameter settings, rely on empirical rules, and generalize poorly. In electricity consumption scenarios with diverse appliance signatures, multi-scale temporal patterns, and overlapping operational behaviors, these limitations are particularly pronounced, underscoring the need for an automated, adaptive segmentation framework.

To evaluate segmentation performance, researchers across diverse domains have developed domain-specific assessment methods. For instance, boundary-distance-sensitive scoring functions have been designed to address the sensitivity of semantic segmentation evaluation to minor discrepancies [10]; tolerance-based precision and recall are used to assess human motion segmentation [11]; the Time2State framework employs the Adjusted Rand Index and Normalized Mutual Information (NMI) for unsupervised state inference [12]; Partition Distortion and Estimation Rate quantify operating-mode boundary detection in radar pulse sequences [13]; and multiple overlap-based metrics, including the Dice score, RMSE, and Mean Error, have been applied to biomedical signal segmentation [14]. Bayesian optimization has also been applied to hyperparameter tuning in related energy analytics tasks [24-26]. Recent surveys have highlighted the pitfalls of relying on single metrics for segmentation and classification tasks [27-29]. This diversity of evaluation approaches underscores the need for a unified, multidimensional evaluation framework tailored to electricity consumption time-series segmentation, in which the hierarchical nature of appliance state transitions demands assessment at both the event and global structural levels.

Existing research on time-series segmentation of electricity consumption has two principal shortcomings. First, segmentation parameters are typically set manually based on domain expertise, making it difficult to adapt to diverse electricity consumption patterns across different electrical environments and monitoring scenarios. Second, existing evaluation methods primarily assess overlap between segmented and ground-truth segments, a metric that is highly sensitive to minor boundary deviations. At the same time, pointwise accuracy metrics often saturate and lack discriminative power in high-quality regions of the parameter space. Together, these limitations hinder the systematic and automated deployment of segmentation-based analysis. This paper addresses these challenges through four contributions.

1) A comprehensive evaluation metric is constructed by integrating the event_$F_1$ score and NMI. The segmentation result is mapped to a binary state sequence of the same length as the original time series, enabling joint assessment along two complementary dimensions: event-detection accuracy and global structural consistency. The event_$F_1$ measures detection precision and recall using a tolerance-matching mechanism that evaluates state-switching events while accommodating practical boundary deviations, thereby capturing local consistency in the segmentation. NMI quantifies agreement between the predicted and ground-truth state sequences, characterizing the global structural consistency of the segmentation. The composite score, obtained through weighted fusion of these two components, effectively overcomes the limitations of conventional pointwise accuracy metrics—namely, their excessive sensitivity to minor boundary disturbances and their insufficient discrimination in high-quality parameter regions—and thus provides a more nuanced and reliable assessment of segmentation performance.
2) BayesSeg, an automatic parameter-optimization framework for segmentation, is built on Bayesian optimization. Using the proposed composite score as the objective function, BayesSeg employs a sequential sampling strategy with surrogate model updates to efficiently search the parameter space for optimal values of the key parameters $\varepsilon$ and $\Delta$. In contrast to an exhaustive grid search, BayesSeg achieves segmentation quality comparable to that of a million-level grid search with only about 100 objective function evaluations, dramatically reducing computational cost. By shifting segmentation parameter configuration from manual, experience-driven tuning to a data-driven optimization paradigm, this framework enables accurate segmentation across diverse datasets and application scenarios without requiring expert intervention.
3) A dual-criteria steady-state discrimination method and an automatic segmentation algorithm featuring sequential extraction and complement-set parsing are proposed. At the implementation level, the dual steady-state discrimination criteria, based on the tail value and mean of the preceding subsequence, are combined with the sequential extraction and complement-set parsing strategy to achieve automatic, precise partitioning of steady-state and transition-state segments in electricity consumption time series, without requiring prior specification of the number of appliance states. This segmentation method operates in concert with the upper-level multidimensional evaluation system and the Bayesian parameter optimization module, forming a unified segmentation–evaluation–optimization framework. The resulting framework provides a transferable, end-to-end solution for time series analysis that can be applied directly to high-resolution, field-collected electrical monitoring data, enabling automated, scalable processing across diverse operational environments.
4) Although BayesSeg is developed and validated in the context of NILM, the proposed framework is inherently domain-agnostic. The segmentation layer relies solely on local statistical properties of the time series—specifically, the tail value and the mean of the preceding subsequence—without requiring domain-specific features or prior knowledge of the number of states. The composite evaluation metric and the Bayesian optimization engine are likewise independent of the signal's physical semantics. BayesSeg can therefore be directly applied to other time-series segmentation tasks in industrial process monitoring, structural health monitoring, biomedical signal analysis, and any domain where distinguishing steady-state regimes from transient events is of interest. The experimental section focuses on electricity consumption data as a representative case study, but the methodological contributions extend well beyond this specific application.

## 2. Model and Methods

During operation, electrical appliances routinely switch between operating states. When an appliance remains in a steady state, the electricity consumption profile on the supply circuit remains relatively stable. Conversely, as the appliance transitions to another state, the power trend exhibits pronounced fluctuations. Subsequent segments within the electricity consumption time series during which the power data remain relatively stable are defined as steady-state segments. Generally, the transition of an appliance from one steady state to another involves a brief transient. The time series segment that characterizes the evolution of electricity consumption over this interval is defined as a transition-state segment.

**Definition 1: Steady-state Segment of Electricity Consumption State Time Series**

Let $S = <s_1, s_2 \dots s_n>$ be the observed time series of data for a specific attribute of the electricity consumption state. Let $es_l = <es_l, es_{l+1} \dots es_{l+m}>$, $es_l \subseteq S$ be the observed data time series segment of the specified attribute over the time period $\Delta_l = (t_l, t_{l+m}]$, where $m$ is the number of observed data points in $\Delta_l$. If $\overline{s} = \frac{1}{m+1}\sum_{i=l}^{l+m} s_i$ and inequality (1) holds, $es_l$ is a steady-state segment of the electricity consumption state time series $S$.

$$\begin{cases} |\overline{s} - s_i| \le \varepsilon \\ |\overline{s} - s_{l+m+1}| > \varepsilon \\ |\overline{s} - s_{l-1}| > \varepsilon \\ |s_i - s_{i+1}| \le \Delta \end{cases} \tag{1}$$

**Definition 2: Transition Segment of Electricity Consumption State Time Series**

Let $ES = <es_1, es_2 \dots es_k>$ be all steady-state segments of the electricity consumption state time series $S$, where $es_i$ $(1 \le i \le k)$ is a steady-state segment in $S$ observed in time period $\left(t_{s_{i,begin}}, t_{s_{i,end}}\right]$ and $t_0 \le t_{s_{i,begin}} \le t_{s_{i,end}} \le t_n$. The observed data segment in $\left(t_{s_{i,begin}}, t_{s_{i,end}}\right]$ is a transition segment of the electricity consumption state time series.

### 2.1. A Segmentation Method based on Tail Value and Mean of the Preceding Subsequence

Figure 1 illustrates the process of segmenting the electricity consumption time series. Algorithm 1 determines whether each observation is locally stable using the steady-state identification criterion, thereby generating candidate steady-state points. Algorithm 2 merges temporally contiguous steady-state points according to Eq. (1) to obtain all steady-state segments within the time series. Based on these steady-state segments, Algorithm 3 extracts all transition segments from the entire electricity consumption state time series, where the sequence fragment between two adjacent steady-state segments constitutes a transition segment. The pseudocode for Algorithms 1, 2, and 3 is given below.

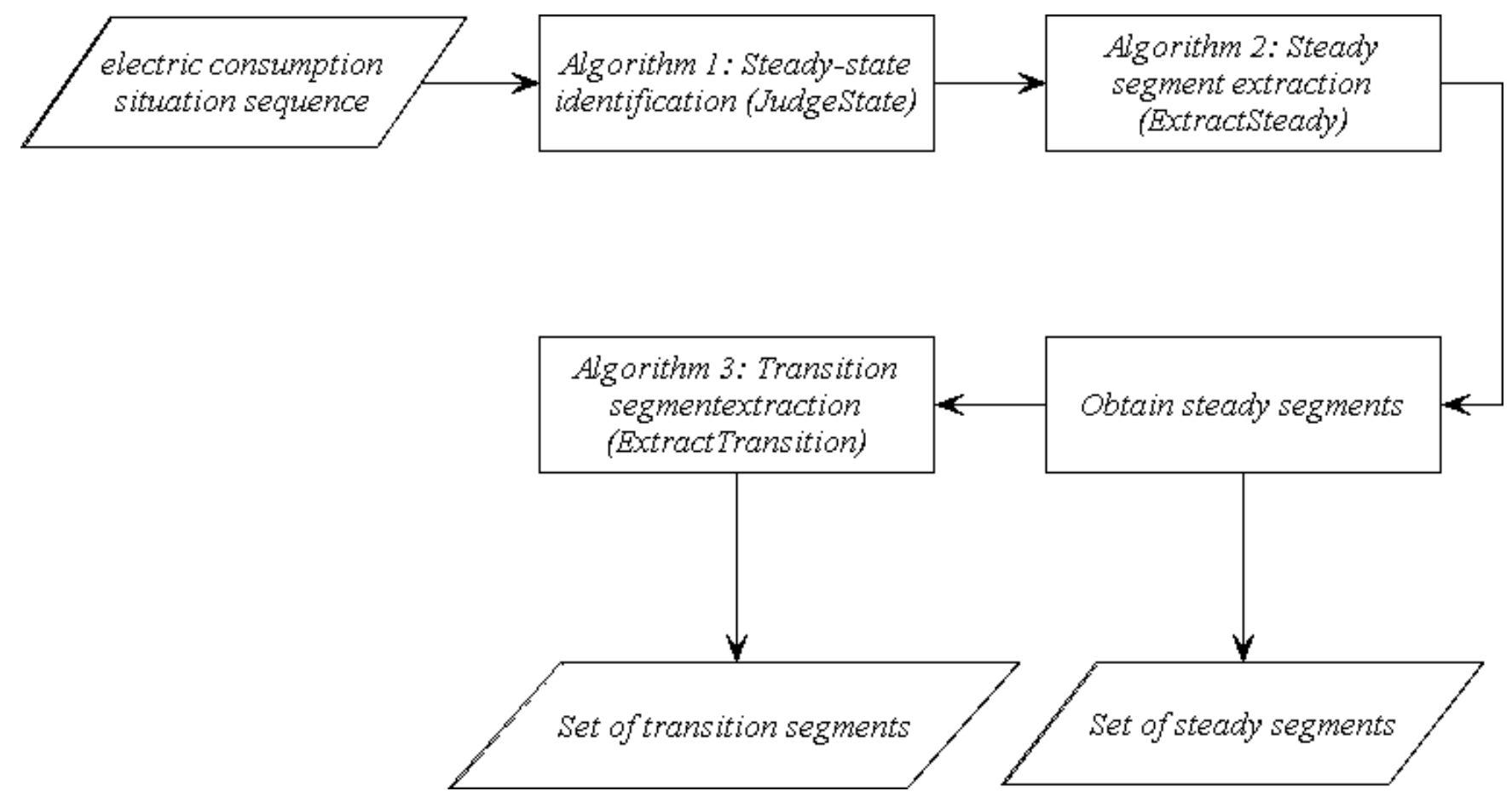

**Figure** 1. Flowchart of the Segmentation

In Algorithm 1, variable es is the current steady-state candidate segment, an ordered set of attribute values that is incorporated into the current steady-state segment and maintained by Algorithm 2. s is the attribute value to be determined, along with thresholds Δ and ε. The variable flag indicates whether s is a data point within a steady-state segment. The expression es[|es|] in step 1 refers to the last data point in the es segment.

Algorithm 1: Steady-State Identification Criterion Based on Local Consistency
Function JudgeState(es, s, Δ, ε)
Input: es, s, Δ and ε
Output: flag

1) T = |s − es[|es|]|
2) M = |s − mean(es)|
3) flag=1;
4) if (T < ε) or (T ≥ ε and M < Δ)
5)     flag=0;
6) return flag

In Algorithm 2, the input current_series is the full electricity consumption time series. The thresholds Δ and ε are defined by inequality (1). The output of Algorithm 2 is a steady-state segment.

```
Algorithm 2: Sequential Extraction Algorithm for Steady-State Segments
Function ExtractSteadySegments(current_series, Δ, ε)
Input: current_series, Δ, ε
Output: steady_segments;
1)    es =φ; steady_segments=φ;
2)    for each s in current_series
3)        if es == φ
4)            es = {s};
5)        else
6)            flag = JudgeState(es, s, Δ, ε);
7)            if flag == 0 then
8)                es = es∪{s};
9)            else
10)               if |es| ≥ 2
11)                   steady_segments = steady_segments  ∪{ es };
12)                   es = {s};
13)               else
14)                   es =φ
15)               end if
16)           end if
17)       end if
18)   end for
19)   return steady_segments
```

In Algorithm 3, current_series represents the full time series of electricity consumption. steady_segments are the output of Algorithm 2. headLoc(seg(head)) and tailLoc(seg(tail)) are used in step 5 of Algorithm 3 to locate the positions of the first and last data points within the segment seg in the time series.

```
Algorithm 3: Extraction Method for Transition-State Segment Sets
Function ExtractTransitionSegments(current_series, steady_segments)
Input: current_series; steady_segments
Output: transition_segments
1)    seg = φ// The steady segment currently being traversed
2)    start = 0; end =0; // Start and end indices of the current steady segment
3)    last_end = 0; // The index immediately after the tail of the previous steady segment
4)    for each seg in steady_segments do
5)        (start, end) ←(headLoc(seg(head)) tailLoc(seg(tail)))
6)        if start > last_end
          / *The current steady segment starts beyond last_end, indicating an unpartitioned region in
             between */
7)            Append the subsequence corresponding to the index interval [last_end, start − 1]
              to transition_segments;
8)            last_end = end + 1;
```

9) end if
10) end for
11) return transition_segments

### 2.2 Evaluation of Electricity Consumption State Time-Series Segmentation

Let $S = \langle s_1, s_2 \ldots s_n \rangle$ be the time series of observations recorded over the time interval $(t_0, t_n]$ on a power line, where the observed physical quantities include current, voltage, power, and others. Let $t_0, t_n \in R$ with $t_0 > 0, t_n > 0$ and $t_0 < t_n$. $b(s_t, t) \in \{0,1\}$ is the binary state label function, where $b(s_t, t) = 1$ indicates that the electricity consumption data at time $t$ is in the transient state, and $b(s_t, t) = 0$ indicates that it is in the steady state. Let $B = \langle b(s_1, t_1), b(s_2, t_2), \ldots, b(s_n, t_n) \rangle$ be the true state sequence corresponding to the actual electricity consumption time series $X$. Let $\hat{B} = \langle \hat{b}(s_1, t_1), \hat{b}(s_2, t_2), \ldots, \hat{b}(s_n, t_n) \rangle$ be the predicted state segmentation result obtained via an electricity consumption time-series segmentation method.

$$Precision = \frac{\sum_{t=1}^{n}\left(b(x_t,t)=1 \wedge \hat{b}(x_t,t)=1\right)}{\sum_{t=1}^{n}\left(b(x_t,t)=1 \wedge \hat{b}(x_t,t)=1\right) + \sum_{t=1}^{n}\left(b(x_t,t)=0 \wedge \hat{b}(x_t,t)=1\right)} \tag{2}$$

$$Recall = \frac{\sum_{t=1}^{n}\left(b(x_t,t)=1 \wedge \hat{b}(x_t,t)=1\right)}{\sum_{t=1}^{n}\left(b(x_t,t)=1 \wedge \hat{b}(x_t,t)=1\right) + \sum_{t=1}^{n}\left(b(x_t,t)=1 \wedge \hat{b}(x_t,t)=0\right)} \tag{3}$$

$$F_1 = \frac{2 \cdot Precsion \cdot Recall}{Precsion + Recal} \tag{4}$$

Given the ground-truth state sequence B for the electricity consumption time series X, a segmentation method can be evaluated by comparing, for each observed data point, whether the ground-truth state matches the identified state. This is a binary classification problem. Typically, the Precision, Recall, and F1 score for evaluating a state segmentation result can be calculated using the following equations (2), (3), and (4), respectively.

Let $G = \{g_1, g_2, ..., g_n\}$ be the set of indices corresponding to real events in the user's electricity consumption time series, where ngt is the total number of events, and let $S = \{s_1, s_2, ..., s_m\}$ be the set of steady-state segments produced by the segmentation algorithm. For each real event $g_i \in G$, if there exists a steady-state segment $s_j \in S$ such that start($g_i$) + $\tau \leq$ start($s_j$) and end($s_j$) $\leq$ end($g_i$) – $\tau$, where $\tau$ is the tolerance parameter, start($g_i$) is the start index of $g_i$, end($g_i$) is the end index of $g_i$, start($s_j$) is the start index of $s_j$, and end($s_j$) is the end index of $s_j$, then $s_j$ is considered a successful match for $g_i$. Here, the indices refer to positions in the electricity consumption time series. This one-to-one matching process ensures that each real event is paired with at most one segment and each segment is paired with at most one event. Precision is the ratio of successfully matched events to the total number of matched segments.

In contrast, recall is the ratio of successfully matched events to the total number of ground-truth events. The event_$F_1$ is computed as the harmonic mean of precision and recall.

While the event_F1 measures the degree of matching accuracy between the identified results and the ground truth, emphasizing local consistency within the sequences, the NMI also quantifies the similarity between the overall structural patterns of the true and predicted state sequences. Specifically, for an electricity consumption time series, if B is the true state sequence and is the predicted state sequence, the NMI can be computed using Eq. (5).

$$NMI\left(B,\hat{B}\right)=\frac{2\cdot I\left(B,\hat{B}\right)}{H\left(B\right)+H\left(\hat{B}\right)} \tag{5}$$

$$H\left(B\right)=-\sum_{b\in\{0,1\}}p\left(b\right)\cdot\log p\left(b\right) \tag{6}$$

$$H\left(\hat{B}\right)=-\sum_{\hat{b}\in\{0,1\}}p\left(\hat{b}\right)\cdot\log p\left(\hat{b}\right) \tag{7}$$

$$I\left(B,\hat{B}\right)=\sum_{b\in\{0,1\}}\sum_{\hat{b}\in\{0,1\}}p\left(b,\hat{b}\right)\log\left(\frac{p\left(b,\hat{b}\right)}{p\left(b\right)p\left(\hat{b}\right)}\right) \tag{8}$$

$$y=\alpha\cdot F1+\left(1-\alpha\right)\cdot NMI \tag{9}$$

Compared with the $F_1$ score, which focuses on the consistency of local time-series labels, the NMI emphasizes the global consistency of time-series labels. By combining the $F_1$ score and the NMI, consistency between the ground truth and the identified time series labels can be assessed at both the local and global levels. Eq. (9) defines a composite score that incorporates both the $F_1$ score and the *NMI*; it captures the accuracy of state identification from a local perspective while measuring the consistency of the overall structure from an information-theoretic perspective.

## 3. Segmentation Parameter Optimization Based on Bayesian Optimization

To achieve adaptive segmentation of user electricity consumption state time series, a Bayesian optimization-based segmentation framework, BayesSeg, is presented in Figure 2. As shown in Figure 2, BayesSeg comprises a segmentation module, an evaluation module, and a parameter-optimization module. Specifically, the segmentation module segments the electricity consumption state time series based on the variation pattern of the tail value of the preceding subsequence and the mean value as defined in Eq. (1); the evaluation module evaluates the segmentation results using the composite score metric; and the parameter-optimization module employs Bayesian optimization to assess the segmentation results, using the composite score as the objective function, thereby determining the optimal segmentation parameters.

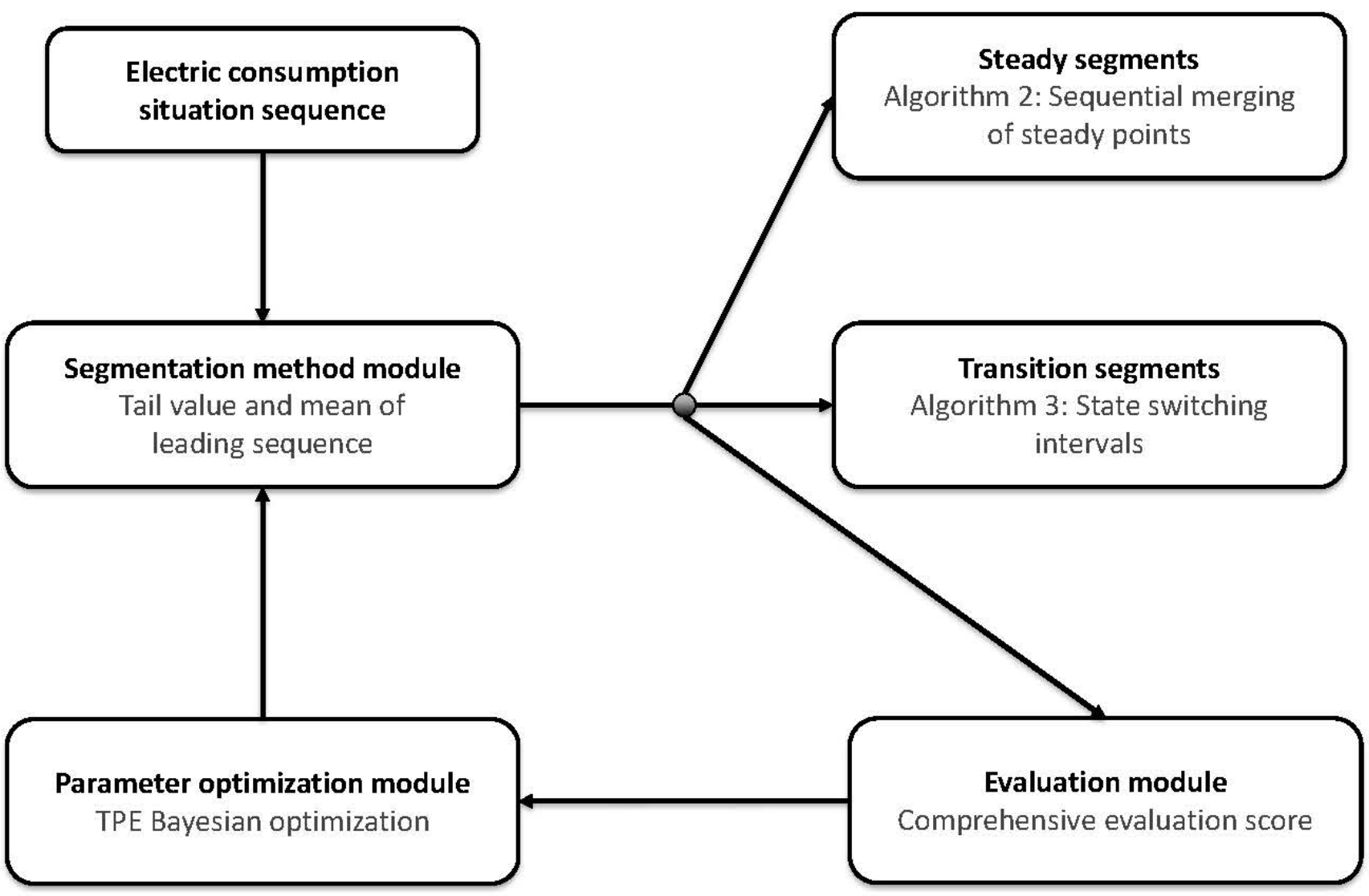

**Figure 2.** BayesSeg Segmentation Framework

ε and Δ are two key parameters in the steady-state segment identification method, which is based on the tail value of the preceding subsequence and its mean, as defined in Eq. (1). Variations in their values will significantly affect the accuracy of the user's electricity consumption time-series segmentation. To quickly obtain optimal ε and Δ, Figure 3 presents a Bayesian optimization framework for user electricity consumption time-series data.

Intuitively, to obtain the optimal ε-Δ pair, one could iterate over all possible ε and Δ values, segment the electricity consumption time series for each combination, and select the combination with the highest composite score. However, a grid search over a large-scale parameter space significantly increases computational overhead. To address this, this paper introduces a Bayesian optimization strategy that efficiently explores the parameter space by constructing a surrogate model. This approach ensures segmentation quality while significantly reducing search time. Figure 3 illustrates the process of searching for the optimal parameter combination using Bayesian optimization.

The Bayesian optimization algorithm approximates the current objective function using the Tree-structured Parzen Estimator (TPE) model to evaluate results and, based on these evaluations, infers the next parameter combination most likely to improve the objective. After each evaluation, the TPE model is updated with the new results, gradually converging toward the optimal solution. When using the Bayesian optimization algorithm to search for the optimal parameter combination for the electricity consumption state-time-series segmentation method, the objective function is given by Eq. (9).

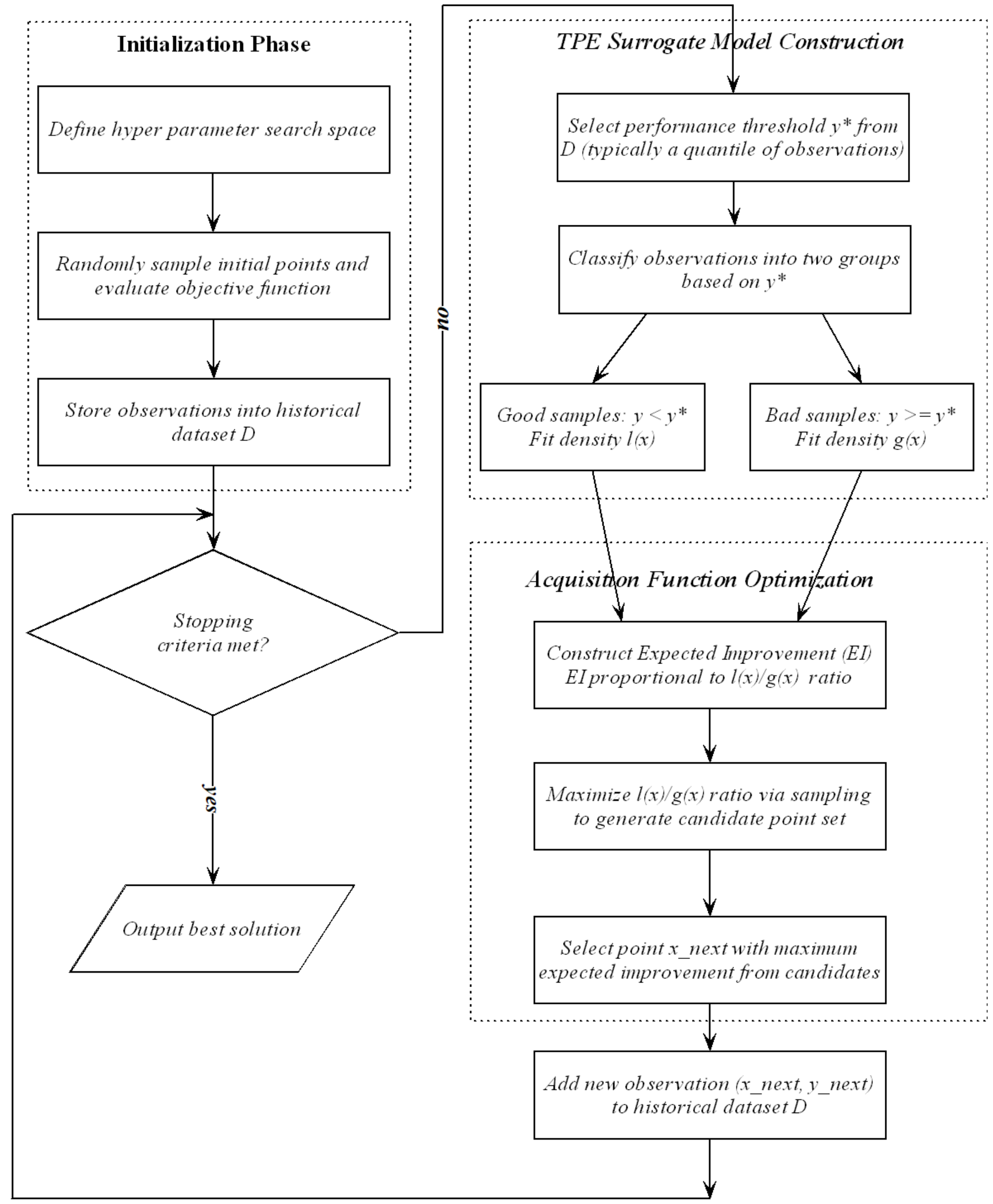

**Figure 3.** Framework of Segmentation Parameter Optimization

To segment an electrical state time series, a candidate parameter pair ($\varepsilon$, $\Delta$) is a 2-dimensional vector $x = (x_1, x_2)^T \in R^2$. A single trial is the process of applying a candidate parameter vector to time-series segmentation, then evaluating the resulting segments using a target function. Let $X_n = \begin{bmatrix} x_{1,1} & x_{1,2} \\ x_{2,1} & x_{2,2} \\ \cdots & \cdots \\ x_{n,1} & x_{n,2} \end{bmatrix}$ be the complete set of candidate parameter vectors from all $n$ historical trials, and let $Y_n = \begin{bmatrix} y_1 \\ y_2 \\ \cdots \\ y_n \end{bmatrix}$ be the corresponding vector of target function values for $X$. The TPE is used to approximate the original objective function. First, the historical parameters are partitioned according to the target function scores defined in Eq. (9). Usually, the top 10% by score are considered good parameters. In contrast, the remaining parameters are considered bad. After that, according to Eq. (11), probability density functions $l$(x) and $g$(x) are estimated over the good and bad parameter subsets, respectively. $\arg\frac{l(x)}{g(x)}$ is calculated as the TPE prediction result. Using the acquisition function, the point $x_{next}$ that maximizes it is determined from the TPE prediction results. $x_{next}$ is then used to segment the user's electricity consumption time series, and the composite score of the segmentation results is used to update the historical parameter set, thereby completing one trial. After each trial, the TPE model is updated with the new observations, and the acquisition function is re-evaluated to guide the next trial. Upon completion of all trials, the parameter combination achieving the highest score is selected as the optimal configuration.

$$\begin{cases} X_{good} = \{x_i \mid y \geq y^*\} \\ X_{bad} = \{x_i \mid y < y^*\} \end{cases} \tag{10}$$

$$\begin{cases} l(x) = p(x \mid y \geq y^*) = \prod_{j=1}^{2} l_j(x_j) \\ g(x) = p(x \mid y < y^*) = \prod_{j=1}^{2} g_j(x_j) \end{cases} \tag{11}$$

$$\begin{cases} l_j(x_j) = \frac{1}{|X_{good}|} \sum_{i \in X_{good}} K_h(x_j - x_{j,i}) \\ g_j(x_j) = \frac{1}{|X_{bad}|} \sum_{i \in X_{bad}} K_h(x_j - x_{j,i}) \end{cases} \tag{12}$$

$y^*$ in Eq. (10) is a threshold, typically chosen as a quantile of the observed objective values $Y_n$. In Eq. (11), $l_j(x_j)$ and $g_j$(xj) are the kernel density estimates for the $j$th parameter dimension for the good and bad parameter subsets, respectively, and $l$(x) and $g$(x) represent the estimated probability densities of the good and bad parameter vectors, respectively. $l_j(x_j)$ and $g_j(x_j)$ are calculated according to Eq. (12). In Eq. (12), $K_h$ is the kernel function with bandwidth $h$, and $x_{j,i}$ is the value of the $i$th sample in the $j$th dimension. Usually, the Gaussian function is the kernel function.

## 4. Experiment Results and Analysis

### 4.1. Experimental Setup

SustDataED2 [15], the dataset used in this experiment, was collected from a real-world residential scenario involving a three-person household in Portugal over a continuous 96-day period. It contains power measurements and power-state transition labels for 18 commonly used household appliances, recorded at a sampling frequency of 0.5 Hz. In the experiments, the monitored power data and corresponding power-state transition labels for the Philips television were used to evaluate the performance of the BayesSeg time-series segmentation model.

To evaluate the effectiveness of Bayesian optimization in determining optimal parameter values for the segmentation model based on the tail and mean of the preceding subsequence, the experiments employed both grid search and Bayesian optimization to identify optimal values of $\varepsilon$ and $\Delta$. The search space for both $\varepsilon$ and $\Delta$ was set to [0 10] ×[0 10]. For grid search, the step size for both parameters was 0.01, yielding 1,002,001 parameter combinations. For Bayesian optimization, the maximum number of iterations was set to 100, and the quantile threshold for distinguishing 'good' from 'bad' parameter configurations was set to 10%. The tolerance parameter $\tau$ was assigned values of 0, 1, 2, 3, 4, 5, and 6. The composite score defined in Eq. (9) was computed using two approaches. The first approach, designated as composite, was calculated from the event_$F_1$ and the NMI value at tolerance $\tau$. The second approach, designated as composite_point, was calculated from the point-level $F_1$ score ($F_1$) and the NMI value at tolerance $\tau$. $\alpha$, the weighting coefficient, was set to 0.5 for all composite score computations.

### 4.2. Effectiveness of the Bayesian Optimization Method

The experiments first employed a grid search strategy, in which the parameters $\varepsilon$ and $\Delta$ each ranged from 0 to 10 in increments of 0.01, resulting in an exhaustive search over 1,002,001 parameter combinations to identify the optimal ($\varepsilon$, $\Delta$) pair. Table 1 presents the maximum values of five evaluation metrics, including $F_1$, event_$F_1$, NMI, composite, and composite_point, along with their corresponding optimal parameter combinations, obtained using the segmentation method based on the tail value and mean of the preceding subsequence, under seven tolerance parameter settings ($\tau$ = 0, 1, 2, 3, 4, 5, 6).

As shown in Table 1, the optimal parameter combinations and the corresponding maxima of the $F_1$ score and NMI are identical across all values of $\tau$. Specifically, $F_1$ attains its maximum of 0.7083 at $\varepsilon$ = 0.26 and $\Delta$ = 1.85, and NMI reaches its maximum of 0.6129 at the same parameter combination. This consistency indicates that, as $\tau$-independent evaluation metrics, $F_1$ and NMI respond to variations in segmentation parameters in a highly aligned manner, with both favoring smaller values of $\varepsilon$ and $\Delta$, which correspond to a fine-grained segmentation strategy that is more sensitive to power fluctuations. A statistical analysis of all 1,002,001 parameter combinations also shows that the coefficients of variation (CV = standard deviation/mean) for $F_1$ and NMI are 58.35% and 58.71%, respectively. Both metrics therefore have strong discriminative power across the parameter space and can effectively guide parameter selection.

Table 1 also shows that event_$F_1$ increases monotonically with $\tau$, rising from 0.9677 at $\tau$ = 0 to 0.9840 at $\tau \geq 3$, where it plateaus. The corresponding optimal parameters shift from $\varepsilon$ = 7.62 and $\Delta$ = 8.36 to $\varepsilon$ = 7.61 and $\Delta$ = 1.07, in marked contrast to the behavior of $F_1$ and NMI. The optimal parameters for event_$F_1$ lie in a region of relatively large $\varepsilon$ and $\Delta$, indicating that event-level detection favors a coarse-grained segmentation strategy with greater tolerance to power variations. This strategy yields fewer steady-state segments with clearer boundaries, thereby achieving higher event detection rates under the tolerance-based matching mechanism. Even at $\tau = 0$ (strict pointwise matching), event_$F_1$ remains at 0.9677. It rises to 0.9840 for $\tau \geq 3$, showing that the tolerance window absorbs the impact of boundary localization errors on event matching.

**Table 1.** Effectiveness of the segmentation method based on the tail value and mean of the preceding subsequence

| | | τ = 0 | τ = 1 | τ = 2 | τ = 3 | τ = 4 | τ = 5 | τ = 6 |
|---|---|---|---|---|---|---|---|---|
| $F_1$ | max | 0.708295 | 0.708295 | 0.708295 | 0.708295 | 0.708295 | 0.708295 | 0.708295 |
| | ε | 0.26 | 0.26 | 0.26 | 0.26 | 0.26 | 0.26 | 0.26 |
| | Δ | 1.85 | 1.85 | 1.85 | 1.85 | 1.85 | 1.85 | 1.85 |
| event_$F_1$ | max | 0.967742 | 0.978723 | 0.978723 | 0.983957 | 0.983957 | 0.983957 | 0.983957 |
| | ε | 7.62 | 7.1 | 7.1 | 7.61 | 7.61 | 7.61 | 7.61 |
| | Δ | 8.36 | 7.73 | 7.73 | 1.07 | 1.07 | 1.07 | 1.07 |
| NMI | max | 0.612892 | 0.612892 | 0.612892 | 0.612892 | 0.612892 | 0.612892 | 0.612892 |
| | ε | 0.26 | 0.26 | 0.26 | 0.26 | 0.26 | 0.26 | 0.26 |
| | Δ | 1.85 | 1.85 | 1.85 | 1.85 | 1.85 | 1.85 | 1.85 |
| composite | max | 0.575879 | 0.712327 | 0.717455 | 0.717455 | 0.717455 | 0.717455 | 0.717455 |
| | ε | 6.58 | 4.2 | 4.2 | 4.2 | 4.2 | 4.2 | 4.2 |
| | Δ | 0 | 2.07 | 2.07 | 2.07 | 2.07 | 2.07 | 2.07 |
| composite_point | max | 0.660593 | 0.660593 | 0.660593 | 0.660593 | 0.660593 | 0.660593 | 0.660593 |
| | ε | 0.26 | 0.26 | 0.26 | 0.26 | 0.26 | 0.26 | 0.26 |
| | Δ | 1.85 | 1.85 | 1.85 | 1.85 | 1.85 | 1.85 | 1.85 |

In Table 1, the optimal parameter combinations for composite and composite_point differ substantially, revealing a fundamental divergence between the two evaluation perspectives in parameter selection. composite_point remains stable at ($\varepsilon = 0.26$, $\Delta = 1.85$) across all values of $\tau$, which exactly matches the optimal parameters for $F_1$ and NMI, since composite_point = $\alpha \times F_1 + (1\text{-}\alpha) \times$ NMI and neither $F_1$ nor NMI varies with $\tau$. In contrast, composite = $\alpha \times$ event_$F_1 + (1\text{-}\alpha) \times$ NMI, and for $\tau \geq 1$, composite stabilizes at ($\varepsilon = 4.20$, $\Delta = 2.07$), with its optimal parameters shifting markedly toward the moderate-ε region, where a better trade-off is achieved between event detection capability and global structural consistency. Note that at $\tau = 0$, the optimal parameter combination for the composite is ($\varepsilon = 6.58$, $\Delta = 0$). A statistical analysis of all parameter combinations shows that 17,075 combinations yield the identical maximum composite value of 0.5759 (accounting for 1.7% of the parameter space), indicating that $\Delta = 0$ is merely a boundary point among tied optima rather than a unique optimum. The strict matching condition at $\tau = 0$ significantly degrades the evaluation metric's discriminative power. For $\tau \geq 1$, the number of combinations that achieve the optimal composite value converges to 214, reducing degeneracy in the parameter space.

Figure 4 shows the distributions of point-level $F_1$ scores and composite scores across the ε-Δ parameter space at $\tau = 4$, visualized with heatmaps and 3D surfaces. As shown in Figures 4(a) and 4(b), high $F_1$ scores cluster in the lower-left corner of the parameter space, within a narrow range where both ε and Δ are small. The heatmaps display a distinct L-shaped pattern. When ε exceeds approximately 1 or Δ exceeds approximately 4, the $F_1$ score drops rapidly from around 0.7 to near zero, and the corresponding 3D surface forms a steep, narrow ridge. This highly localized distribution indicates that $F_1$ is extremely sensitive to the segmentation parameters. A smaller ε implies a lower tolerance for deviations in the tail of the power time series, producing more fine-grained segments and higher precision and recall in pointwise matching evaluation. However, once ε or Δ becomes too large, misalignment of segment boundaries intensifies, leading to a sharp deterioration in point-level matching performance.

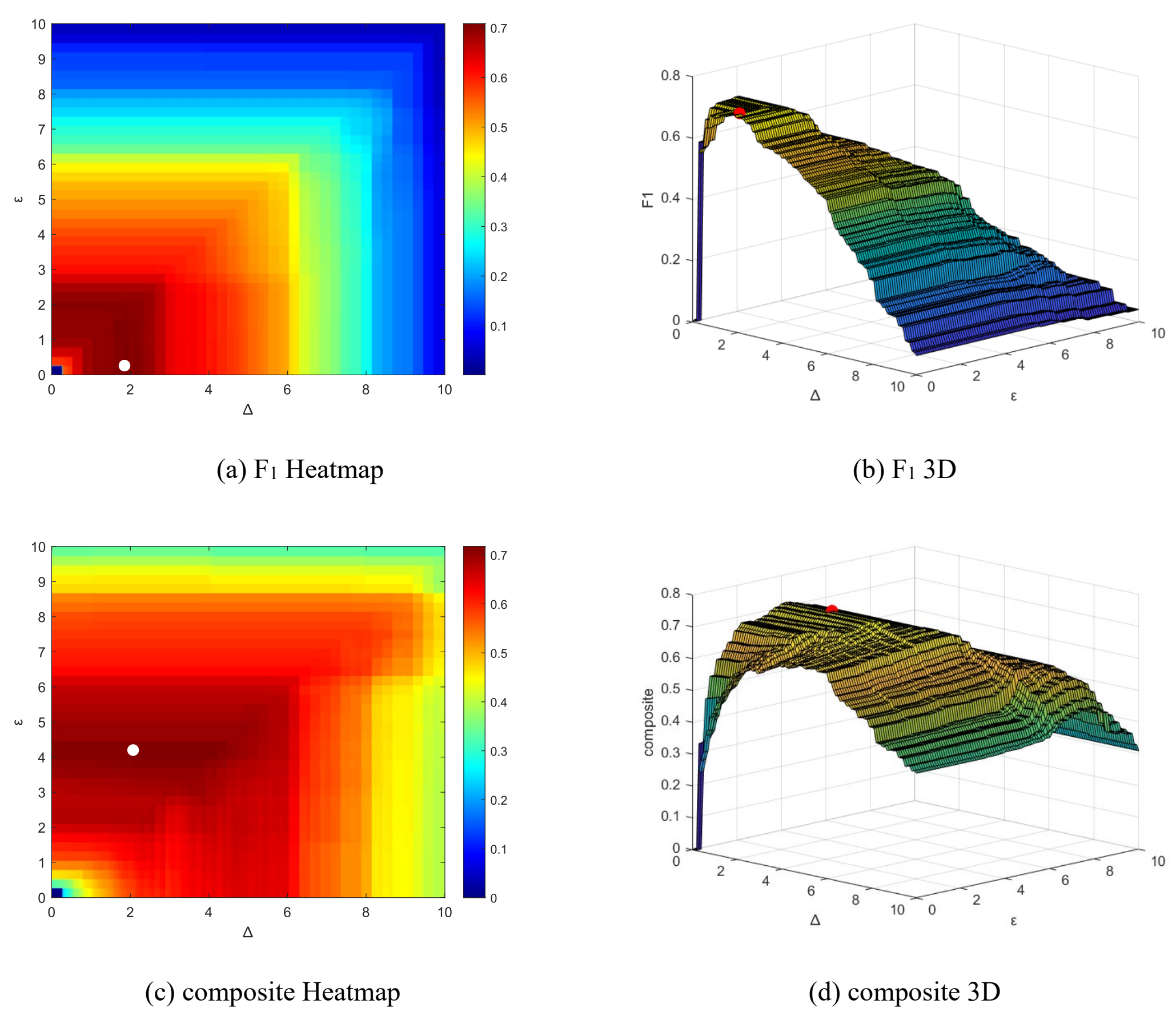

(a) $F_1$ Heatmap (b) $F_1$ 3D

(c) composite Heatmap (d) composite 3D

**Figure 4**. Sensitivity of the segmentation method based on the tail value and mean of the preceding subsequence to the ε-Δ parameters (τ = 4)

In contrast, the composite score distributions in Figures 4(c) and 4(d) show distinct topographical features. The high-value regions (composite ≥ 0.65) form a broad, plateau-like band extending along the Δ direction, spanning ε from approximately 3 to 6 and Δ from approximately 0 to 5. Across all 1,002,001 parameter combinations, over 30% achieve more than 90% of the maximum composite value, indicating that the composite score is robust to parameter selection. The optimal point for the composite score is (ε=4.20, Δ=2.07), with a value of 0.7175. This parameter combination neither tends toward extremely small values, such as the $F_1$ optimal point (ε=0.26, Δ=1.85), nor toward extremely large values, such as the event_$F_1$ optimal point (ε=7.61, Δ=1.07); instead, it achieves an effective trade-off between point-level accuracy and event detection capability.

Both heatmaps also show that the composite score approaches zero at the origin of the parameter space, where ε = 0 and Δ = 0. This result suggests that an overly fine segmentation strategy also fails to yield a satisfactory comprehensive evaluation. Although segment boundaries are dense in this scenario, the excessive fragmentation of steady-state segments severely compromises global structural consistency, as measured by NMI. The 3D surfaces confirm this observation. The $F_1$ surface exhibits a nearly vertical, cliff-like decline from the peak to the valley, whereas the composite surface shows gently rolling hills. This morphological difference intuitively reflects the fundamental distinction in parameter sensitivity between point-level and composite evaluations.

**Table 2.** Comparison of optimal results

| τ | method | ε | Δ | $F_1$ | event_$F_1$ | NMI | composite | composite_point |
|---|---|---|---|---|---|---|---|---|
| 0 | GS | 6.58 | 0 | 0.309711 | 0.885417 | 0.266342 | 0.575879 | 0.288027 |
| | BO | 6.82896 | 6.22119 | 0.305263 | 0.890052 | 0.262448 | 0.57625 | 0.283856 |
| 1 | GS | 4.2 | 2.07 | 0.560268 | 0.933333 | 0.49132 | 0.712327 | 0.525794 |
| | BO | 3.96697 | 3.79923 | 0.565072 | 0.923858 | 0.49584 | 0.709849 | 0.530456 |
| 2 | GS | 4.2 | 2.07 | 0.560268 | 0.94359 | 0.49132 | 0.717455 | 0.525794 |
| | BO | 3.96697 | 3.79923 | 0.565072 | 0.93401 | 0.49584 | 0.714925 | 0.530456 |
| 3 | GS | 4.2 | 2.07 | 0.560268 | 0.94359 | 0.49132 | 0.717455 | 0.525794 |
| | BO | 3.96697 | 3.79923 | 0.565072 | 0.93401 | 0.49584 | 0.714925 | 0.530456 |
| 4 | GS | 4.2 | 2.07 | 0.560268 | 0.94359 | 0.49132 | 0.717455 | 0.525794 |
| | BO | 3.96697 | 3.79923 | 0.565072 | 0.93401 | 0.49584 | 0.714925 | 0.530456 |
| 5 | GS | 4.2 | 2.07 | 0.560268 | 0.94359 | 0.49132 | 0.717455 | 0.525794 |
| | BO | 3.96697 | 3.79923 | 0.565072 | 0.93401 | 0.49584 | 0.714925 | 0.530456 |
| 6 | GS | 4.2 | 2.07 | 0.560268 | 0.94359 | 0.49132 | 0.717455 | 0.525794 |
| | BO | 3.96697 | 3.79923 | 0.565072 | 0.93401 | 0.49584 | 0.714925 | 0.530456 |

Tables 2 and 3 list two methods in the method column: Grid Search (GS) and Bayesian Optimization (BO). Table 2 presents the optimal parameter combinations for each method, along with the corresponding scores for the evaluation metrics, including $F_1$, event_$F_1$, NMI, composite, and composite_point. Table 3 presents the minimum, maximum, mean, and variance of the time per search iteration, as well as the total time required to complete the optimization.

As shown in Table 2, the composite scores from the two methods are nearly identical across all 7 tolerance parameter sets, with a maximum difference of only 0.00253. For τ ranging from 2 to 6, GS yields a score of 0.717455 and BO yields 0.714925, corresponding to a relative difference of 0.35%. When $\tau \geq 2$, the composite score stabilizes, indicating that the tolerance window is sufficient to cover the matching range for event detection. These results demonstrate that BO can locate an optimal parameter region nearly equivalent to that of a grid search while using only 100 objective function evaluations, which account for 0.01% of the 1,002,001 evaluations required by GS. When $\tau \geq 1$, the optimal parameters for GS remain stable at $\varepsilon$=4.20 and $\Delta$=2.07, while those found by BO remain stable at $\varepsilon$=3.97 and $\Delta$=3.80. Although these two sets of parameters are far apart in the parameter space, both lie within adjacent plateau regions, and the difference in composite score is merely 0.35%. This also supports the broad plateau of high composite values observed in Figure 4, where a vast number of near-optimal solutions exist in the parameter space, and BO can discover solutions of comparable quality without an exhaustive search. Notably, when GS found the maximum composite value of 0.575879 at $\Delta = 0$ and $\tau = 0$, a closer look at the full data shows that in the optimal ε region from 6.58 to 6.82, all 683 values of Δ ranging from 0 to 6.82 yield an identical composite score, and Δ = 0 is therefore merely a boundary point among the co-optimal solutions ordered lexicographically. In contrast, BO identified an interior point with ε = 6.82896 and Δ = 6.22119, achieving a composite score of 0.57625, which is slightly higher than the GS result by 0.000371. This outcome occurs primarily because BO searches in continuous space rather than discrete grids, enabling it to explore higher-quality points within the region of co-optimal solutions. The two methods also show complementary differences in individual metrics. For τ ranging from 1 to 6, BO achieves slightly higher $F_1$ and NMI scores than GS, yet yields marginally lower event_$F_1$ scores. This pattern reveals the trade-offs among sub-metrics within the plateau regions where each method's optimal parameters reside. Given that the composite weighted metric values remain similar, minute shifts in the parameter space can lead to reciprocal increases and decreases across the sub-metrics.

### 4.3. Efficiency of Bayesian Optimization

Table 3 presents the time consumption data for two methods used to search for the optimal parameter combination. As shown in Table 3, there is a significant disparity in computational efficiency between the two methods. Across all τ values, the total time for GS to complete all 1,002,001 evaluation points exceeded 5,260 seconds, whereas BO finished within 1 second. This implies that, while ensuring segmentation quality, the BO method incurs negligible time overhead compared to the GS method. Furthermore, a comparison of the minimum, maximum, mean, and variance of the time consumed per evaluation reveals a severely right-skewed distribution for GS single evaluations; this suggests that a few parameter combinations triggered computationally intensive boundary conditions in GS.

**Table 3.** Comparison of Optimization Time

| **τ** | **method** | **optimal parameter combination** | | **time cost per parameter combination (ms)** | | | | **total time cost (s)** |
|---|---|---|---|---|---|---|---|---|
| | | **ε** | **Δ** | **min** | **max** | **mean** | **std** | |
| 0 | GS | 6.58 | 0 | 4.421 | 1912.503 | 5.291954333 | 19.04950218 | 5302.54353 |
| | BO | 6.828959 | 6.221187 | 5 | 14 | 9.21 | 1.327981684 | 0.921 |
| 1 | GS | 4.2 | 2.07 | 4.413 | 729.468 | 5.258701786 | 18.62560211 | 5269.22445 |
| | BO | 3.96697 | 3.799225 | 5 | 14 | 9.3 | 1.352140092 | 0.93 |
| 2 | GS | 4.2 | 2.07 | 4.423 | 760.414 | 5.288041275 | 19.30384161 | 5298.62265 |
| | BO | 3.96697 | 3.799225 | 5 | 14 | 9.29 | 1.312680964 | 0.929 |
| 3 | GS | 4.2 | 2.07 | 4.414 | 728.991 | 5.252258065 | 18.55179091 | 5262.76783 |
| | BO | 3.96697 | 3.799225 | 5 | 14 | 9.22 | 1.33014771 | 0.922 |
| 4 | GS | 4.2 | 2.07 | 4.419 | 752.484 | 5.283607788 | 19.04811459 | 5294.18029 |
| | BO | 3.96697 | 3.799225 | 5 | 15 | 9.28 | 1.385932206 | 0.928 |
| 5 | GS | 4.2 | 2.07 | 4.418 | 740.675 | 5.274087042 | 18.72912793 | 5284.64049 |
| | BO | 3.96697 | 3.799225 | 5 | 14 | 9.35 | 1.373449539 | 0.935 |
| 6 | GS | 4.2 | 2.07 | 4.417 | 3145.402 | 5.2845752 | 18.93284071 | 5295.14964 |
| | BO | 3.96697 | 3.799225 | 5 | 15 | 9.27 | 1.391569131 | 0.927 |

By contrast, the per-iteration time for BO is highly uniform. The speedup ratios between the two methods exceed 5,600 for all τ values, with a mean of approximately 5,701. This indicates that in practical applications of piecewise parameter optimization, BO reduces search cost from hours to seconds, improving time efficiency by nearly 4 orders of magnitude. Meanwhile, search quality degrades by less than 0.35%. Regarding computational resource utilization, although the time cost of a single BO iteration is approximately 9.3 ms, which is slightly higher than the average 5.3 ms per GS evaluation, the overall computational load is reduced because the total number of evaluations has decreased from millions to hundreds.

**Table 4.** Exploring Efficiency of the Bayesian Optimization Method

| τ | ε | Δ | composite | mean(composite) | std(composite) | first_hit | total_time(sec) |
|---|---|---|---|---|---|---|---|
| 0 | 6.828959 | 6.221187 | 0.57625 | 0.52124745 | 0.069982323 | 34 | 0.921 |
| 1 | 3.96697 | 3.799225 | 0.709849 | 0.64761754 | 0.094631605 | 29 | 0.93 |
| 2 | 3.96697 | 3.799225 | 0.714925 | 0.65482567 | 0.093500543 | 29 | 0.929 |
| 3 | 3.96697 | 3.799225 | 0.714925 | 0.65619052 | 0.09267299 | 29 | 0.922 |
| 4 | 3.96697 | 3.799225 | 0.714925 | 0.65622257 | 0.092648797 | 29 | 0.928 |
| 5 | 3.96697 | 3.799225 | 0.714925 | 0.65622257 | 0.092648797 | 29 | 0.935 |
| 6 | 3.96697 | 3.799225 | 0.714925 | 0.65622257 | 0.092648797 | 29 | 0.927 |

The exploration efficiency data in Table 4 provide more detail on the BO method's convergence during parameter optimization. Across all seven experiment sets ($\tau = 0$ to 6), the BO method reached the global optimal composite value within 29 to 34 iterations out of a maximum of 100. Specifically, the optimal value of 0.5763 was found at iteration 34 for $\tau = 0$, while the optimal value of 0.7149 was found at iteration 29 for $\tau = 1$ to 6. If 99% of the optimal composite value is adopted as the convergence criterion, $\tau = 0$ requires only 12 iterations (0.082 s), $\tau = 1$ requires 20 iterations (0.155 s), and $\tau = 2$-6 require 28 iterations (≈0.23 s). This finding suggests that in engineering deployments, the number of BO iterations can be reduced from the preset value of 100 to approximately 30 while still attaining more than 99% of the optimal value, thereby compressing the total search time to roughly 0.23 seconds. The case of $\tau = 0$ exhibits the fastest convergence rate, reaching the 99% threshold in merely 12 iterations. This rapid convergence can be attributed to the distinctive structure of the parameter space under the $\tau = 0$ condition. The strict point-by-point matching criterion at $\tau = 0$ confines the high-value regions of the composite metric to an extremely narrow area, covering only 1.7% of the parameter space. The TPE sampler can therefore quickly eliminate vast regions of low quality and focus on the confined high-value band in a limited number of iterations. In contrast, for $\tau \geq 1$, the high-value regions expand into a broad plateau, requiring the TPE sampler to run more iterations to locate the optimal point within this expansive region precisely. Across all τ values, the mean of the composite metric ranges from 0.52 to 0.66, with a standard deviation of approximately 0.07 to 0.095, indicating that the TPE sampler explored extensive quality gradients across the parameter space over 100 iterations rather than performing intensive sampling solely in local regions. This global exploration capability is the key advantage of Bayesian optimization over random search.

### 4.4. Comparison with Other Optimization Methods

The preceding experiments (Tables 2–4) have validated the efficiency advantage of Bayesian Optimization (BO) over Grid Search (GS). GS is an exhaustive strategy without adaptive search capability. Whether Bayesian optimization still retains a clear advantage over heuristic methods such as particle swarm optimization (PSO) and genetic algorithm (GA) remains an open question for the BayesSeg optimization module. If an alternative method can achieve search quality comparable to that of BO with a shorter runtime under a limited evaluation budget, a simpler optimizer could be adopted within the framework. To investigate this, three baselines—PSO, GA, and random search (RS)—are introduced for a multi-metric comparison with BO under unified conditions.

The experimental configurations for the four methods are as follows. Bayesian optimization uses the TPE sampler within the Optuna framework. Particle swarm optimization uses the Clerc constriction factor model with an inertia weight of $\omega = 0.7298$, learning factors $c1 = c2 = 1.49445$, and a swarm population size of 20. The genetic algorithm uses real-valued encoding, simulated binary crossover (SBX) with a distribution index of 20, and polynomial mutation with a distribution index of 20, with a population size of 10. Random search draws uniform random samples within the search space as a baseline. A swarm size of 20 for PSO and a population size of 10 for GA ensure that the 100-evaluation budget corresponds to 5 and 10 generations, respectively. For the two-dimensional optimization problem involving ($\varepsilon$, $\Delta$), 10–20 candidate solutions per generation provide sufficient spatial coverage. All four methods share the same parameter search space ($\varepsilon \in [0,10]$, $\Delta \in [0,10]$, both continuous) and evaluation budget (100 objective function evaluations). A paired experimental design is adopted: the four methods use the same set of 50 random seeds (42–91), and under a given seed, different methods share the same initial sampling points, yielding paired observations that control variance due to randomness and enhance statistical power. The total computational workload amounts to 4 (methods) × 7 ($\tau$ values) × 50 (seeds) = 1400 independent optimization runs. Statistical testing employs the Friedman test for overall differences, and the Wilcoxon signed-rank test for pairwise comparisons, with Bonferroni correction applied ($\alpha_{adj} = 0.05/3 \approx 0.0167$); the effect size is quantified using paired Cohen's d.

Table 5 presents the comparison results for the four methods—BO, PSO, GA, and RS—grouped by the tolerance parameter $\tau$ (0–6). The columns are defined as follows. The composite column reports the mean and standard deviation of the optimal composite scores across 50 independent runs; the superscript ** indicates that the difference between BO and the given method is statistically significant after Bonferroni correction (paired Wilcoxon signed-rank test, $p<0.0167$). The first_hit column denotes the number of evaluation rounds required to first reach 99% of the GS global optimum (mean ± standard deviation), with the proportion of runs that successfully attained this threshold across 50 trials shown in parentheses, serving as a measure of convergence reliability. The time column records the total runtime for 100 evaluations (mean ± standard deviation, in milliseconds).

As shown in Table 5, BO achieves the highest composite mean across all values of $\tau$. For $\tau=4$, BO attains a composite score of 0.7165 ± 0.0019, deviating from the GS global optimum of 0.7175 by only 0.14%, indicating that 100 evaluations are sufficient to closely approximate the result of an exhaustive search. PSO follows closely (0.7157 ± 0.0018), and its gap from BO is not significant after Bonferroni correction ($p > 0.0167$ for $\tau \geq 2$; Cohen's d≈0.27~0.30, representing a small effect). In contrast, GA (0.7115 ± 0.0072) and RS (0.7130 ± 0.0039) are significantly inferior to BO ($p = 0.0000$; Cohen's d = 0.74 and 0.88, respectively, indicating medium to near-large effects), and both exhibit larger standard deviations, reflecting poorer search stability. Regarding convergence speed, BO yields the smallest

first_hit value (16.6 ± 7.6 at τ = 4), with all 50 runs reaching the target threshold (50/50); PSO requires 22.4 ± 12.3 evaluations (50/50); RS (32.6 ± 24.1, 43/50) and GA (36.0 ± 30.2, 39/50) converge more slowly, with 14% and 22% of their runs, respectively, failing to reach the threshold within 100 evaluations. Although GA theoretically possesses global search capability, with a population of only 10 individuals evolved over 10 generations, the selection pressure is insufficient for the search to fully unfold, resulting in the slowest convergence in practice. In terms of computational efficiency, PSO, GA, and RS exhibit comparable runtimes (280–308 ms), while BO requires approximately 530–550 ms, roughly 1.8 times that of the other methods. The additional overhead primarily stems from updating the TPE probabilistic model; nevertheless, the total runtime remains well under one second, posing no bottleneck in practical use. Overall, BO leads in both solution quality and convergence speed, and the additional computational overhead is acceptable, making it a suitable choice for parameter tuning in BayesSeg.

**Table 5.** Multi-tolerance parameter comparison of four optimization methods (50 independent runs, evaluation budget of 100)

| τ | method | composite | first_hit | time (ms) |
|---|---|---|---|---|
| 0 | **BO** | 0.5760±0.0002 | 12.8±6.3 (50/50) | 539±97 |
| | PSO | 0.5759±0.0004** | 20.5±15.3 (50/50) | 294±97 |
| | GA | 0.5737±0.0039** | 27.1±27.1 (43/50) | 278±4 |
| | RS | 0.5752±0.0020** | 29.0±18.2 (49/50) | 305±96 |
| 1 | **BO** | 0.7116±0.0020 | 15.6±6.8 (50/50) | 550±117 |
| | PSO | 0.7104±0.0016** | 24.1±14.5 (50/50) | 299±96 |
| | GA | 0.7064±0.0069** | 35.4±30.8 (37/50) | 279±4 |
| | RS | 0.7081±0.0034** | 32.6±24.1 (43/50) | 304±94 |
| 2 | **BO** | 0.7164±0.0019 | 15.9±7.0 (50/50) | 530±68 |
| | PSO | 0.7156±0.0018 | 23.0±13.4 (50/50) | 308±115 |
| | GA | 0.7110±0.0080** | 33.4±28.6 (37/50) | 280±5 |
| | RS | 0.7130±0.0039** | 32.6±24.1 (43/50) | 304±96 |
| 3 | **BO** | 0.7165±0.0019 | 16.6±7.6 (50/50) | 534±68 |
| | PSO | 0.7157±0.0018 | 22.4±12.3 (50/50) | 308±115 |
| | GA | 0.7115±0.0072** | 36.0±30.2 (39/50) | 280±5 |
| | RS | 0.7130±0.0039** | 32.6±24.1 (43/50) | 304±93 |
| 4 | **BO** | 0.7165±0.0019 | 16.6±7.6 (50/50) | 532±69 |
| | PSO | 0.7157±0.0018 | 22.4±12.3 (50/50) | 308±115 |
| | GA | 0.7115±0.0072** | 36.0±30.2 (39/50) | 280±5 |
| | RS | 0.7130±0.0039** | 32.6±24.1 (43/50) | 303±94 |
| 5 | **BO** | 0.7165±0.0019 | 16.6±7.6 (50/50) | 530±68 |
| | PSO | 0.7157±0.0018 | 22.4±12.3 (50/50) | 308±115 |
| | GA | 0.7115±0.0072** | 36.0±30.2 (39/50) | 280±5 |
| | RS | 0.7130±0.0039** | 32.6±24.1 (43/50) | 303±94 |
| 6 | **BO** | 0.7165±0.0019 | 16.6±7.6 (50/50) | 531±69 |
| | PSO | 0.7157±0.0018 | 22.4±12.3 (50/50) | 308±115 |
|  | GA | $0.7115 \pm 0.0072^{**}$ | 36.0±30.2 (39/50) | 280±5 |
|  | RS | $0.7130 \pm 0.0039^{**}$ | 32.6±24.1 (43/50) | 304±95 |

Note: composite = 0.5 × event_$F_1$ + 0.5 × NMI. first_hit is the number of function evaluations required to first reach 99% of the ground-truth (GS) optimum composite value; values in parentheses indicate the number of runs (out of 50) that reached this threshold. Statistical comparisons of composite values were performed using two-tailed paired Wilcoxon signed-rank tests. The significance threshold was adjusted to $\alpha = 0.0167$ via Bonferroni correction for three pairwise comparisons (BO vs. PSO, GA, and RS). Methods marked with ** are significantly outperformed by BO ($p < 0.0167$); unmarked methods show no significant difference. All methods were run with the same 50 random seeds to enable paired comparisons and control for stochastic variability.

### 4.5. Discussion

The BayesSeg framework was systematically evaluated through five sets of experiments (Tables 1–5) and a parameter-sensitivity visualization (Figure 4), focusing on validity, performance comparisons, and efficiency. The key findings are as follows.

First, the composite score, which integrates event_$F_1$ and NMI, effectively overcomes the limited discriminative power of traditional accuracy metrics for assessing segmentation parameters. The full grid search results in Table 1 show that all five evaluation metrics are highly discriminative across a parameter range spanning three orders of magnitude, with $F_1$ varying from 0.001 to 0.708 and event_$F_1$ from 0 to 0.984. The coefficients of variation for $F_1$ and NMI approached 59%, and only about 8% of parameter combinations achieved 90% of their respective maxima. Figure 4 provides an intuitive visualization of the fundamental difference in parameter sensitivity between point-level and composite evaluations. The L-shaped cliff in the point-level $F_1$ heatmap stands in marked contrast to the broad plateau of the composite surface. Point-level $F_1$ is highly sensitive to parameter variations, with high values confined to a narrow range where $\varepsilon<1$ and $\Delta<4$. In contrast, the composite score maintains near-optimal performance across a substantially wider region, specifically where $\varepsilon$ lies approximately between 3 and 6 and $\Delta$ lies approximately between 0 and 5. This differential response pattern enables the joint evaluation system to guide parameter selection at different granularities, thereby avoiding potential bias introduced by reliance on any single metric. The tolerance parameter $\tau$ enables event-level evaluation to flexibly adapt to varying precision requirements for boundary localization across different application scenarios. As $\tau$ increased from 0 to 3, event_$F_1$ rose monotonically from 0.9677 to 0.9840, indicating that the tolerance window effectively accommodated minor deviations in boundary positioning.

Second, Bayesian optimization achieved high search efficiency in optimizing segmentation parameters for the segmentation method based on the tail value and mean of the preceding subsequence. As shown in Tables 2–4, the TPE sampler required only 100 objective function evaluations to identify optimal parameters, and the resulting composite score differed by no more than 0.35% from that obtained via grid search. Regarding convergence behavior, Bayesian optimization first reached the global optimum in approximately 29-34 iterations. When adopting 99% of the optimal value as the convergence criterion, it converged in merely 12-28 iterations, corresponding to a search time of 0.08–0.23 seconds. This efficiency advantage enables BayesSeg to reduce parameter optimization time from approximately 88 minutes required by grid search to under 1 second in practical deployments, achieving a speedup of 5,701× with a search-quality degradation of less than 0.35%. Notably, Bayesian optimization operates over continuous spaces and is therefore not constrained by discrete grid boundaries. For $\tau = 0$, the grid search was trapped at a boundary degeneracy point where $\Delta = 0$, which ranked first in lexicographical order among 683 tied optima. In contrast, Bayesian optimization successfully identified a slightly superior interior point, yielding a composite score of 0.5763 compared to 0.5759. This shows the advantage of continuous-space search in avoiding discretization artifacts. Furthermore, the optimal parameters identified by Bayesian optimization across different $\tau$ values ($\varepsilon = 3.97$, $\Delta = 3.80$) and those found by grid search ($\varepsilon = 4.20$, $\Delta = 2.07$) are far apart in the parameter space, yet their composite scores are nearly identical. This observation indirectly corroborates the broad plateau characteristic of the composite metric revealed in Figure 4: a large number of approximately equivalent optimal solutions exist in the parameter space, and Bayesian optimization can efficiently identify solutions of comparable quality without having to traverse the entire parameter space.

Third, the relationship between optimization performance and the evaluation budget is worth discussing. The core principle of TPE is to build a probabilistic surrogate model from existing evaluations and use it to select the parameter configuration with the highest information gain as the next evaluation point. In each evaluation round, the model is updated, and the search shifts toward regions with high composite values. Consequently, virtually no evaluation rounds are "wasted" over 100 evaluations. In contrast, PSO and GA both maintain a population of candidate solutions. Under the constraint of 100 evaluations, PSO (population size 20) undergoes only 5 generations, while GA (population size 10) undergoes 10 generations. Both the selection pressure across generations and the transmission of genetic information are limited, making it difficult for the global search advantages of swarm intelligence to take effect—a point corroborated by GA's largest first_hit standard deviation and lowest success rate. For NILM and broader sequence analysis tasks, when a single evaluation of the objective function is expensive (e.g., large-scale sequence segmentation or cross-validated training), the evaluation budget is typically limited. In such cases, surrogate-model-based Bayesian optimization is more suitable than population-based methods. If the budget were substantially increased (e.g., to 500 evaluations or more), GA might narrow the gap through more thorough evolution, but this lies beyond the scope of the scenarios BayesSeg currently addresses.

## 5. Conclusion and Future Work

The time-series segmentation method based on tail values and the means of preceding subsequences is an effective approach for segmenting user electricity consumption time series. Effective segmentation of these state sequences significantly reduces the complexity of annotating user electricity consumption states. To evaluate the segmentation results, this study constructs a composite score that combines the event_$F_1$ and NMI. This metric captures local precision in state identification while measuring the consistency of the overall structure from an information-theoretic perspective. Experimental results demonstrate that the composite score has higher sensitivity and discriminative power. Furthermore, this study proposes the BayesSeg framework, which uses Bayesian optimization to rapidly identify the optimal parameter combination for the segmentation method based on the tail value and mean of the preceding subsequence. This framework enables automatic configuration of segmentation strategies, reduces reliance on manual expertise, and enhances the method's generalization and applicability across diverse electricity consumption scenarios.

The experiments were validated solely on 0.5 Hz sampling data from a single household and a single appliance (Philips TV); therefore, the BayesSeg framework's generalization across diverse electricity consumption scenarios—such as concurrent operation of multiple appliances, varying sampling rates, and different household structures—remains to be verified. Furthermore, BayesSeg currently uses a fixed composite weight (0.5×event_F1 + 0.5×NMI) to evaluate segmentation performance, without experimentally validating the sensitivity of the composite metric to the weight parameter α or analyzing how different weight allocations affect the selection of optimal parameters. However, practical applications may require varying emphasis on event detection rates versus global structural consistency, depending on the scenario. Consequently, adaptive weight adjustment or exploration of the Pareto Front for the segmentation method remains to be explored. In particular, exploring the Pareto Front shifts the paradigm from "making weight decisions for the user" to "presenting all optimal trade-offs to the user," enhancing the framework's adaptability across different application scenarios. This would improve the robustness of electricity consumption state time-series segmentation methods in complex environments.

## Acknowledgements

The authors acknowledge support from the Discipline (Major) Top-notch Talent Academic Funding Project of Anhui Provincial University and College under Grants gxbjZD2021067 and gxyq2022030; the Innovative Leading Talents Project of the Anhui Provincial Special Support Program under Grant [2022]21; the Key Project of Natural Science Research of Universities of Anhui Province under Grant 2024AH050246; and the open foundation of the Anhui Province Key Laboratory of Intelligent Building & Building Energy Saving under Grant IBES2020KF09.

## Abbreviations

The following abbreviations are used in this manuscript:

| | |
|---|---|
| NILM | Non-Intrusive Load Monitoring |
| event_$F_1$ | event-level $F_1$ score |
| BO | Bayesian Optimization |
| GS | Grid Search |
| NMI | Normalized Mutual Information |
| TPE | Tree-structured Parzen Estimator |
| PSO | Particle Swarm Optimization |
| GA | Genetic Algorithm |
| RS | Random Search |
| SBX | Simulated Binary Crossover |